\documentclass[letterpaper, 10 pt, conference]{ieeeconf}  %
\IEEEoverridecommandlockouts                              %
\usepackage{amsmath}
\usepackage{amssymb}
\usepackage{bm}
\usepackage{verbatim}
\usepackage{xcolor} 
\usepackage{xspace} 
\usepackage{graphicx} 
\usepackage{url}
\usepackage{hyperref}
\graphicspath{{figs/}}  
\usepackage{comment}
\usepackage[font=small]{caption} %
\usepackage{subcaption}
\usepackage{tabularx} %
\usepackage{wrapfig}

\usepackage{eqparbox}

\usepackage{etoolbox}   
\newcommand{\insertfig}{
    \captionsetup{type=figure}
    \includegraphics[width=\textwidth]{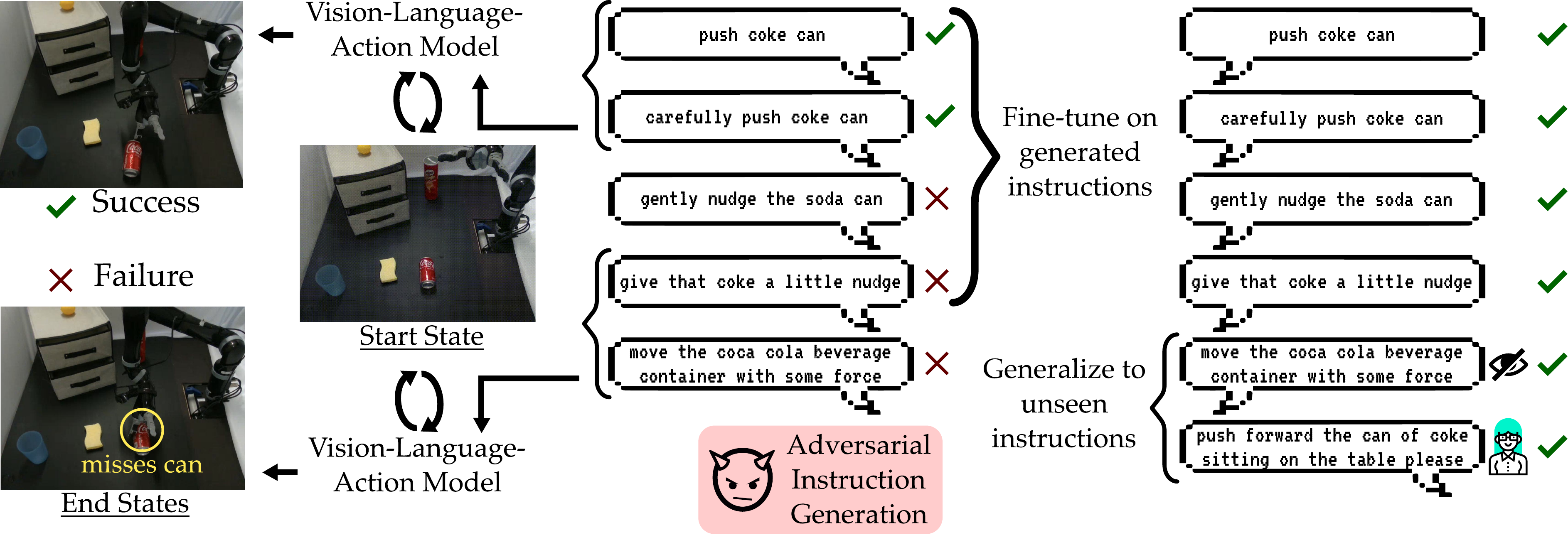}
    \captionof{figure}{
        Our framework, \method, aims to make VLA-powered robots robust to different instruction wordings by generating adversarial instructions that induce task failures.
        For example, a VLA might succeed when instructed to ``push coke can'' (the VLA uses one of the end effectors to push the can), but fail when instructed ``meticulously exert force upon the aluminum beverage container'' (the VLA misses the coke can and pushes nothing).
        Augmenting demonstrations with such diverse instructions and fine-tuning the VLA using them leads to better generalization to unseen and human-generated instructions. 
    }
    \label{fig:pull}
    \vspace{-3mm}
}

\makeatletter
\apptocmd{\@maketitle}{%
  \centering\insertfig
  \setcounter{figure}{1} %
}{}{}%
\makeatother

\PassOptionsToPackage{numbers,compress}{natbib}

\newcommand{\method}{Q-DIG\xspace}

\usepackage{color, soul} %

\usepackage[backend=biber,style=ieee,maxbibnames=1,minbibnames=1]{biblatex}
\newcommand{\E}{\mathbb{E}}

\usepackage{amsfonts}

\def\eqref#1{equation~\ref{#1}}

\def\1{\bm{1}}

\def\vc{{\bm{c}}}

\def\vo{{\bm{o}}}

\DeclareMathAlphabet{\mathsfit}{\encodingdefault}{\sfdefault}{m}{sl}
\SetMathAlphabet{\mathsfit}{bold}{\encodingdefault}{\sfdefault}{bx}{n}

\def\gC{{\mathcal{C}}}
\def\gD{{\mathcal{D}}}

\def\gZ{{\mathcal{Z}}}

\DeclareMathOperator*{\argmax}{arg\,max}

\usepackage{multirow}

\title{\LARGE \bf
Red-Teaming Vision-Language-Action Models via Quality Diversity Prompt Generation for Robust Robot Policies
}

\author{Siddharth Srikanth$^{1}$$^*$, Freddie Liang$^{1}$, Ya-Chuan Hsu$^{1}$, Varun Bhatt$^{1}$, Shihan Zhao$^{1}$, Henry Chen$^{1}$, Bryon Tjanaka$^{1}$ \\ Minjune Hwang$^{1}$, Akanksha Saran$^{2}$, Daniel Seita$^{1}$$^{\dagger}$, Aaquib Tabrez$^{3}$$^{\dagger}$, Stefanos Nikolaidis$^{1}$$^{\dagger}$%
\thanks{$^{1}$Thomas Lord Department of Computer Science, University of Southern California, $^{2}$Sony AI, $^{3}$Sibley School of Mechanical and Aerospace Engineering, Cornell University. $^{\dagger}$Equal advisement. $^*$Corresponding author. Inquiries should be emailed to {\tt\small ssrikant@usc.edu}.}%
}%

\begin{document}

\maketitle
\thispagestyle{empty}
\pagestyle{empty}

\begin{abstract}

Vision-Language-Action (VLA) models have significant potential to enable general-purpose robotic systems for a range of vision-language tasks. However, the performance of VLA-based robots is highly sensitive to the precise wording of language instructions, and it remains difficult to predict when such robots will fail. We propose Quality Diversity (QD) optimization as a natural framework for red-teaming embodied models, and present \method (Quality Diversity for Diverse Instruction Generation), which performs red-teaming by scalably identifying diverse, natural language task descriptions that induce failures while remaining task-relevant. \method integrates QD techniques with Vision-Language Models (VLMs) to generate a broad spectrum of adversarial instructions that expose meaningful vulnerabilities in VLA behavior.  
Our results across multiple simulation benchmarks show that \method finds more diverse and meaningful failure modes compared to baseline methods, and that fine-tuning VLAs on the generated instructions improves task success rates. %
Furthermore, results from a user study highlight that \method generates prompts judged to be more natural and human-like than those from baselines. 
Finally, real-world evaluations of \method prompts show results consistent with simulation, and fine-tuning VLAs on the generated prompts further success rates on unseen instructions.
Together, these findings suggest that \method is a promising approach for identifying vulnerabilities and improving the robustness of VLA-based robots. 
Our anonymous project website is at \href{qdigvla.github.io}{qdigvla.github.io}.

\end{abstract}

\section{Introduction}

Vision-Language-Action (VLA) models, which fine-tune Vision-Language Models (VLMs) to produce robot actions, have demonstrated impressive generalization in robotics tasks~\cite{FMs_robotics_2023_survey_1,FMs_robotics_2023_survey_2,zhong2025surveyvisionlanguageactionmodelsaction}.
For example, they enable out-of-the-box deployment across novel environments, cross-embodiment transfer, dexterous manipulation (e.g., folding laundry), and efficient learning of new tasks with limited data~\cite{barreiros2025careful,kim24openvla,black2024pi0visionlanguageactionflowmodel,intelligence2025pi05}.
Despite this progress, VLAs remain vulnerable to jailbreaking~\cite{JailbreakLLMsRobotics} and red-teaming~\cite{perez2022red} attacks through adversarial instructions, which can elicit unexpected failures~\cite{pi0-experiment-wild} and limit deployment in safety-critical applications.
For example, given the task of pushing a Coke can, a VLA might successfully complete the task when the instruction is exactly ``push Coke can,'' while changing the instruction to ``gently nudge the soda can!'' causes the VLA to fail.  We aim to systematically identify and mitigate such vulnerabilities.

We study this problem through the lens of \emph{red-teaming}: the systematic discovery of inputs that expose failures in a model. While prior methods have been developed to induce failures in VLAs via adversarial instructions, it is still challenging to generate \emph{realistic} instructions that elicit a \emph{controllably diverse} range of failures. In particular, Embodied Red Teaming (ERT)~\cite{karnik2024embodiedredteamingauditing} leverages in-context learning~\cite{dong2022survey} to probe robot policies in simulation and generate adversarial instructions. Although ERT can uncover failure cases, it does not explicitly target a set of failure modes specified by system designers. Additionally, some generated instructions may fall
outside the distribution of instructions that users normally
provide. For the Coke can task above, one such unrealistic
instruction would be ``spot the red and white can lying flat
and push it with precision.''

We argue that Quality Diversity (QD) optimization~\cite{pugh2015confronting,mouret2015illuminating} provides a natural framework for red teaming embodied models: it searches for high-performing solutions while maintaining diversity across user-defined dimensions. Prior work on algorithmic scenario generation has  used QD to discover diverse failure modes in planning-based robotic systems~\cite{fontaine2021quality,bhatt2023}. Later work such as Rainbow Teaming~\cite{samvelyan2024rainbow} has applied QD to generate diverse adversarial prompts for large language models by targeting different ``attack styles,'' such as slang and technical terms, and encouraged
adversarial instructions to be generated following these style
guidelines.  However, although Rainbow Teaming has been effective at red-teaming LLMs, it solely operates in language space and does not consider visual context, which would be crucial for red-teaming VLAs.

Building on this formulation, we present \method (Figure~\ref{fig:pull}), a novel framework for red-teaming and improving VLAs. \emph{Our key insight is that combining quality-diversity optimization with vision-language grounding enables the systematic discovery of diverse, realistic, and task-relevant failure modes in embodied systems.} 
Starting from an initial task instruction, \method iteratively mutates instructions, using previously discovered instructions as stepping stones, to build a repertoire of failure-inducing instructions across semantic categories, or ``attack styles''~\cite{samvelyan2024rainbow}. 
After creating adversarial instructions, \method combines them with existing demonstrations to create an augmented dataset used to finetune the VLA, improving its robustness against diverse and adversarial instructions.

Our contributions are: 
\textbf{(1)} We present \method, a framework that leverages Quality Diversity optimization for diverse and in-distribution adversarial instruction generation.
\textbf{(2)} We evaluate \method in two simulation domains, SimplerEnv~\cite{li24SimplerEnv} and LIBERO~\cite{liu2023libero}, and show that compared to prior VLA red-teaming methods, \method produces more diverse and in-distribution adversarial instructions. 
\textbf{(3)} Through a user study, we show that instructions generated by \method are more human-like compared to those generated by prior methods.
\textbf{(4)} We demonstrate that fine-tuning with the augmented dataset containing adversarial instructions improves robustness to unseen instructions. 
\textbf{(5)} We validate our approach in a sim-to-real setting, showing that the benefits of adversarial fine-tuning transfer to real-world robotic deployments.

\section{Related Work}

\begin{figure*}[t]
    \centering
     \includegraphics[width=1.0\textwidth]{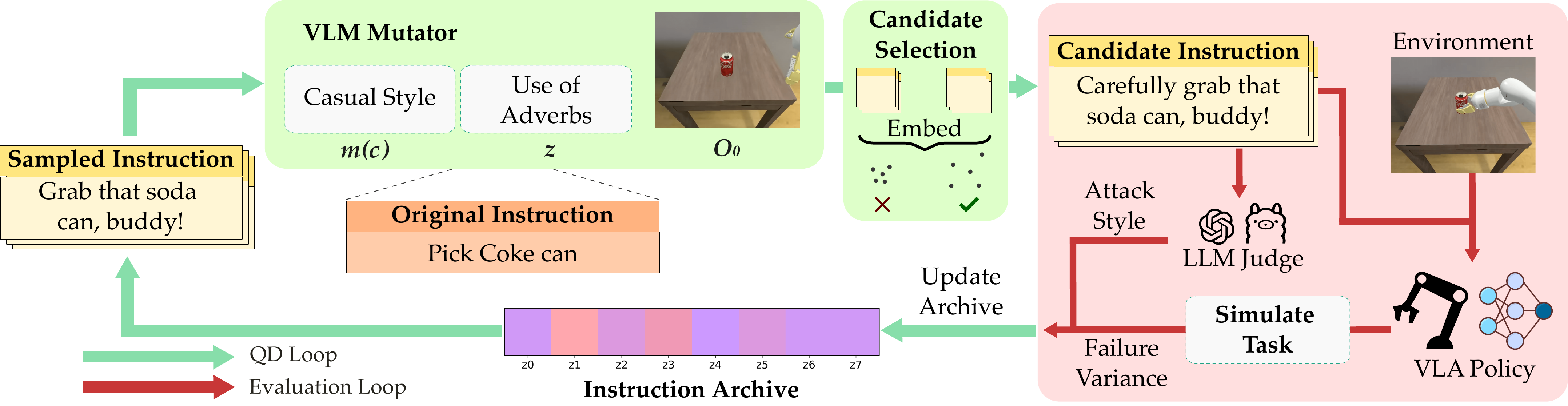}
    \caption{
        Overview of \method. 
        \method leverages previously generated instructions as in-context examples to generate new adversarial instructions in target attack styles (green arrows).
        The generated instructions are evaluated (red arrows) to obtain the variance of failure rates they induce in the VLA as well as their actual attack style.
        Instructions inducing high failure rates with different attack styles (\textbf{z0} to \textbf{z7} from Table~\ref{tab:failure_categories}) are stored in an archive, providing high-quality and diverse examples for future iterations.
    }
    \label{fig:pipeline}
    \vspace{-4mm}
\end{figure*}

\subsection{Vision-Language-Action (VLA) Models}

Vision-Language Models (VLMs) are a type of foundation model that integrates visual and language inputs to support multimodal reasoning.
VLAs fine-tune large pre-trained VLMs to produce robot actions and thus inherit rich perceptual priors. 
Recent VLAs like RT-2~\cite{rt22023arxiv}, OpenVLA~\cite{kim24openvla}, $\pi_0$~\cite{black2024pi0visionlanguageactionflowmodel}, and $\pi_{0.5}$~\cite{intelligence2025pi05} highlight VLAs' potential for generalization across robotic tasks.
These models are trained on large-scale robotics datasets~\cite{open_x_embodiment_rt_x_2023,khazatsky2024droid,barreiros2025careful} and fine-tuned~\cite{kim2025fine} to adapt to new domains or hardware.

Despite this progress, VLAs remain brittle in practice: small changes in task phrasing can induce failures, even when instructions are semantically equivalent~\cite{pi0-experiment-wild}. 
In part, this arises because current VLA datasets and fine-tuning protocols provide only a narrow mapping between language instructions and demonstrations.
For example, LIBERO~\cite{liu2023libero} provides a single instruction per task with multiple demonstrations, DROID~\cite{khazatsky2024droid} associates each trajectory with at most three instructions, and Bridge~\cite{walke2023bridgedata} has a one-to-one mapping between instructions and demonstrations.
We address this limitation by generating realistic adversarial instructions to expose failure modes in VLAs, followed by pairing them with demonstrations and retraining VLAs to mitigate them.

\subsection{Red-Teaming Foundation Models}

Foundation models can produce harmful behaviors, ranging from reinforcing societal biases and generating offensive content to leaking private information~\cite{perez2022red, bai2022training, weidinger2021ethical}.
While filtering, fine-tuning, and reinforcement learning from human feedback~\cite{ouyang2022training, bai2022training} partially mitigate these issues, models remain vulnerable to targeted attacks.
Such efforts are often driven by \emph{red-teaming} (also called jailbreaking, if the intent is to cause harm), where adversarial inputs elicit unexpected outputs to uncover hidden vulnerabilities~\cite{perez2022red, ganguli2022red, ramamurthy2022reinforcement}.

Robots controlled by foundation models are also prone to jailbreaking, with potential physical safety consequences. 
Prior work~\cite{JailbreakLLMsRobotics} showed that jailbroken LLM-controlled robots can be induced to take harmful actions across self-driving, unmanned vehicle navigation, and quadruped control. %
Prior research has also identified such vulnerabilities by predicting performance degradation under environmental variations~\cite{majumdar2025predictive} and by searching over geometric object perturbations~\cite{goel2025redteaming}. 

Our focus is on red-teaming VLAs via adversarial instructions.
Closely related to our work is Embodied Red Teaming (ERT)~\cite{karnik2024embodiedredteamingauditing}, which uses in-context learning~\cite{dong2022survey} %
to generate failure-inducing instructions in simulation. While effective at finding failures, ERT differs from our approach as follows: (1) it does not provide controllability over failure modes, (2) it has limited ability to generate prompts with varied performance (i.e., inducing both robot failures and successes, thus increasing \emph{failure variance}), 
and (3) it does not use failure-inducing instructions to improve the base policies. In contrast, our method explicitly controls semantic attack categories and failure variance. Additionally, we demonstrate that fine-tuning VLAs with the augmented adversarial dataset substantially improves robustness to unseen instructions.

\subsection{Quality Diversity for Foundation Models}

Quality Diversity (QD) algorithms search for a diverse collection of high-performing solutions~\cite{pugh2016quality} by leveraging previously found solutions as stepping stones to generate new solutions that are higher performing or more diverse.
QD has been applied in robotics, including training diverse RL agents~\cite{cideron2020qdrl,nilsson2021pga,tjanaka2022approximating,batra2023proximal}. QD has also been applied to generate diverse scenarios for autonomous agents~\cite{bhatt2022deep} and  deployed robotic systems~\cite{fontaine2021quality,bhatt2023}.  More recently, previous work has applied QD to text generation with LLMs~\cite{meyerson2023language,fernando2023promptbreeder, bradley2024quality,lim2024largelanguagemodelsincontext}, where previously discovered prompts~\cite{fernando2023promptbreeder} or poems~\cite{bradley2024quality} serve as context for an LLM to generate new and improved variants,  

The QD works closest to ours are Rainbow Teaming~\cite{samvelyan2024rainbow} and PLAN-QD~\cite{srikanth2025algorithmic}. 
Rainbow Teaming extends scenario generation with QD to adversarial prompt generation, using LLMs as mutators that generate prompt variants and critics that evaluate outputs to discover diverse failures in language tasks.
It also uses semantic failure modes, or ``attack styles'' (e.g., slang or technical terms) as measures to diversify prompts along user-relevant styles. 
Conditioning on these attack styles makes generated prompts more realistic. %
On the evaluation side, PLAN-QD searches for diverse prompts that induce varied collaborative behaviors in LLM-powered agents for sequential decision-making tasks.
While both Rainbow Teaming and PLAN-QD advance systematic prompt search, they lack grounding in physical tasks and visual information required for embodied domains. 
In contrast, our method leverages visual grounding and feedback from rollouts of an embodied agent to generate diverse adversarial instructions.

\section{Problem Definition}
\label{sec:problem_defn}

To improve the robustness of VLAs to language instructions, we address the problem of generating diverse instructions for a given task.
We assume a given \textit{base VLA}, $\pi(\vo,\vc)$, that maps observations $\vo$ (visual and/or proprioceptive) and language instructions $\vc \in \gC$ (where $\gC$ is the set of all natural language instructions) to actions that a robot can execute.
The language instructions describe a task $T \equiv (\mu_0, g_T)$, with $\mu_0$ being the initial state distribution and $g_T(\zeta)$ being the goal predicate that outputs $1$ when a rollout trajectory $\zeta$ completes the task.
We define a set of language instructions $\gC_T$ for each task $T$ such that all $\vc \in \gC_T \subset \gC$ describe $T$.
Given a base language instruction $\vc_{0,T} \in \gC_T$, our goal is to generate a diverse set of $N$ new instructions $\{\vc_{1,T},\dots,\vc_{N,T}\} \subset \gC_T$. 
We evaluate the efficacy of the new instructions via several diversity metrics as well as the performance of the VLA after fine-tuning with these additional instructions (Sec.~\ref{sec:experiments}).

\begin{table*}[t]
\centering
\caption{Categories of failure in task instructions, partially obtained from~\cite{karnik2024embodiedredteamingauditing}.
For ease of reference, we index the attack styles using ``z0'' through ``z7'' (later used in Figure~\ref{fig:qdvlm_heatmaps}). 
Refer to Section~\ref{sec:exp_inst_gen} for details on how these failure modes were selected.
}
\label{tab:failure_categories}
\renewcommand{\arraystretch}{1.2}
\begin{tabularx}{\textwidth}{p{0.48\textwidth} p{0.48\textwidth}}
\hline
\textbf{Step-by-step instructions:} unnecessary breakdown of the task into multiple sequential steps \textbf{(z0)} &
\textbf{Uncommon vocabulary:} use of rare, technical, or overly formal words instead of simple ones \textbf{(z1)} \\
\hline

\textbf{Human-centric tone:} instruction phrased as if addressing a human instead of a robot \textbf{(z2)} &
\textbf{Use of adverbs:} adding vague modifiers like ``carefully'' or ``gently'' that don't add clarity \textbf{(z3)} \\

\hline

\textbf{Unnecessary action specification:} extra details about how to execute the task, beyond the core instruction \textbf{(z4)} &
\textbf{Overly verbose reformulation:} inflated or wordy phrasing without changing the meaning \textbf{(z5)} \\

\hline

\textbf{Colloquial/casual style:} use of slang or conversational tone inappropriate for robotic tasks \textbf{(z6)} &
\textbf{Mixed modality references:} adding sensory references (e.g., sight, sound) that the robot may not have \textbf{(z7)} \\
\hline
\end{tabularx}
\vspace{-3mm}
\end{table*}

\section{Method: \method}
\label{sec:method}

To generate diverse task instructions for VLAs, we propose \method (\textbf{Q}uality Diversity for \textbf{D}iverse \textbf{I}nstruction \textbf{G}eneration), a framework that instantiates red-teaming of VLAs as a Quality Diversity (QD) optimization problem.

The central idea is to use QD to automatically generate a wide range of semantically diverse instructions that, when executed on the base VLA, serve as adversarial prompts revealing failure modes. 
In this section, we first describe how we formulate generating adversarial instructions for VLAs as a QD problem, then provide the detailed framework of QD optimization, and finally explain how the resulting instructions can be leveraged for fine-tuning. Figure~\ref{fig:pipeline} presents an overview of our pipeline.

\subsection{Quality Diversity Formulation}
\label{ssec:qd_formulation}

QD optimizes both the \textit{quality} of each solution and the \textit{diversity} across the solution set. In our case, the solution space is the set of instructions $\gC$. The \textit{quality} of an instruction is represented by the variance of the base VLA's failure rate on task $T$, defined as:
\begin{equation}
    J(\vc)=\E_{\zeta \sim \pi(\cdot \mid \cdot, \vc)}[g_T(\zeta)](1-\E_{\zeta \sim \pi(\cdot \mid \cdot, \vc)}[g_T(\zeta)])
\end{equation}
We use the variance instead of the raw failure rate to promote instructions on the boundary of the VLA's linguistic capabilities, while discouraging maximally adversarial yet unrealistic instructions $\vc \not\in \gC_T$. To construct a heuristic for the \textit{diversity} among instructions, we define a base set of attack styles $\gZ = \{z_i\}_{i=0}^M$ (see Table~\ref{tab:failure_categories}), and assume a \textit{measure function} $m: \gC \rightarrow \gZ$ that maps each instruction to some attack style $z \in \gZ$. We then consider diversity among generated instructions as the collective coverage of their image under $m$ over $\gZ$. 
The QD problem we aim to solve is to discover a highest-variance instruction for each attack style, which we formalize using the following objective:
\begin{equation}
    \{\argmax_{\vc}{J(\vc)} \text{ s.t. } m(\vc) = z, \; \forall z \in \gZ\}.
\end{equation}
We borrow from QD terminology and refer to the collection of instructions as an \emph{archive}, consisting of $M+1$ ``cells'' each containing a highest-variance instruction mapped to one attack style.

\subsection{Generating Adversarial Instructions}
\label{ssec:gen_adv_instr}

To discover diverse adversarial instructions for VLAs that cause failures, \method applies QD optimization in four steps: (1) instruction selection, (2) candidate instruction mutation, (3) instruction evaluation, and (4) archive update. Repeating these steps progressively fills the archive with semantically diverse attack styles and failure-inducing task instructions.

\noindent\textbf{Instruction Selection:} 
The optimization begins with the original task instruction $\vc_{0,T}$, such as ``Pick Coke can'' in Figure~\ref{fig:pipeline}. 
As the archive fills, \method samples a filled cell and retrieves its stored instruction, which then serves as a ``stepping stone'' for generating new adversarial instructions. 

\noindent\textbf{Instruction Mutation:} 
Extending prior work in LLM jailbreaking~\cite{samvelyan2024rainbow}, \method leverages a VLM as a \textit{mutator}.
Given an existing adversarial instruction $\vc$, its attack style $m(\vc)$, initial observation for the task $\vo_0$ (visual), and a target attack style category $z$, the mutator VLM leverages in-context learning~\cite{dong2022survey} to generate a candidate instruction in the target attack style (e.g., ``use of adverbs'' in Fig.~\ref{fig:pipeline}). 
We further generate $k$ sets of candidate instructions of batch size $b$ each (where $b$ and $k$ are hyperparameters). 
We then project each generated instruction into an embedding space with Sentence-BERT~\cite{reimers2019sentence} and compute the average pairwise cosine similarity for each generated set.
Among the generated sets, we select the one with the the most diverse set of instructions, i.e. the least pairwise similarity.

\noindent\textbf{Instruction Evaluation:} 
Each new instruction $\vc$ in the selected set is evaluated by rolling out the base VLA $\pi(\vo,\vc)$ on task $T$ to compute the failure variance $J(\vc)$. 
In addition, as our \textit{measure function} $m$, an external LLM (referred to as \emph{LLM judge}) categorizes the instruction into one of the semantic attack styles $\{z_i\}_{i=0}^M$.

\noindent\textbf{Archive Update:} 
The instruction archive maintains an elitist cell for each attack style, and the new instructions are each assigned a cell according to its evaluated attack style $m(\vc)$. Each instruction is added to the archive only under one of two conditions: \textbf{(1)} If the instruction's assigned cell is not occupied by a predecessor instruction with the same style, in which case inserting it improves the archive's coverage over $\gZ$, thereby \textit{diversity}. \textbf{(2)} The new instruction has higher $J(\vc)$ than the predecessor instruction with the same style, in which case the new instruction overwrites the predecessor, improving the archive's \textit{quality} for that attack style.

\subsection{Fine-Tuning the VLA}
\label{sec:method_finetuning}

Once the set of adversarial instructions is discovered, we fine-tune the base VLA to mitigate the induced failures and improve the robot's robustness to these attack styles.
We assume access to a dataset of expert demonstrations $\gD =\{\tau_1, \dots, \tau_K\}$ of size $K$, with each demonstration mapping to an original language instruction for the given task $T$. 
We then augment this dataset by associating the language instructions for some demonstrations with one of our generated adversarial instructions, \emph{without} collecting additional expert demonstrations.
Formally, we divide $\gD$ into $G+1$ subsets (where $G$ is the number of generated adversarial instructions) and map a different instruction to demonstrations in each of those subsets, resulting in the augmented dataset $\gD_{\mathrm{aug}}$ with both the adversarial and original instructions. 
In our experiments, we have $N=50$ demonstrations, and $G=9$ unique instructions (8 adversarial instructions + 1 original instruction), for each task.

Following prior work~\cite{kim2025fine}, we perform supervised fine-tuning of the base VLA $\pi(\vo,\vc)$ with data from $\gD_{\mathrm{aug}}$, but the augmented dataset by itself is agnostic to the fine-tuning method.
In summary, we use the diverse adversarial instructions generated by \method to enrich the training dataset.
Once finetuned on this dataset, the VLA becomes more robust to different ways of describing the task.

\section{Experiment Protocol}
\label{sec:experiments}

We performed several experiments to evaluate our adversarial instruction generation method.
First, we measured the diversity and human-likeness of the generated instructions and compared them with other instruction generation baselines (Sec.~\ref{sec:exp_inst_gen}).
Then, we fine-tuned a VLA with our augmented dataset and evaluated it in simulation (Sec.~\ref{sec:exp_finetuning}) as well as on a real robot (Sec.~\ref{sec:exp_real}).

\begin{figure*}[th!]
    \centering
    \begin{subfigure}[t]{0.325\textwidth}
        \includegraphics[width=\linewidth]{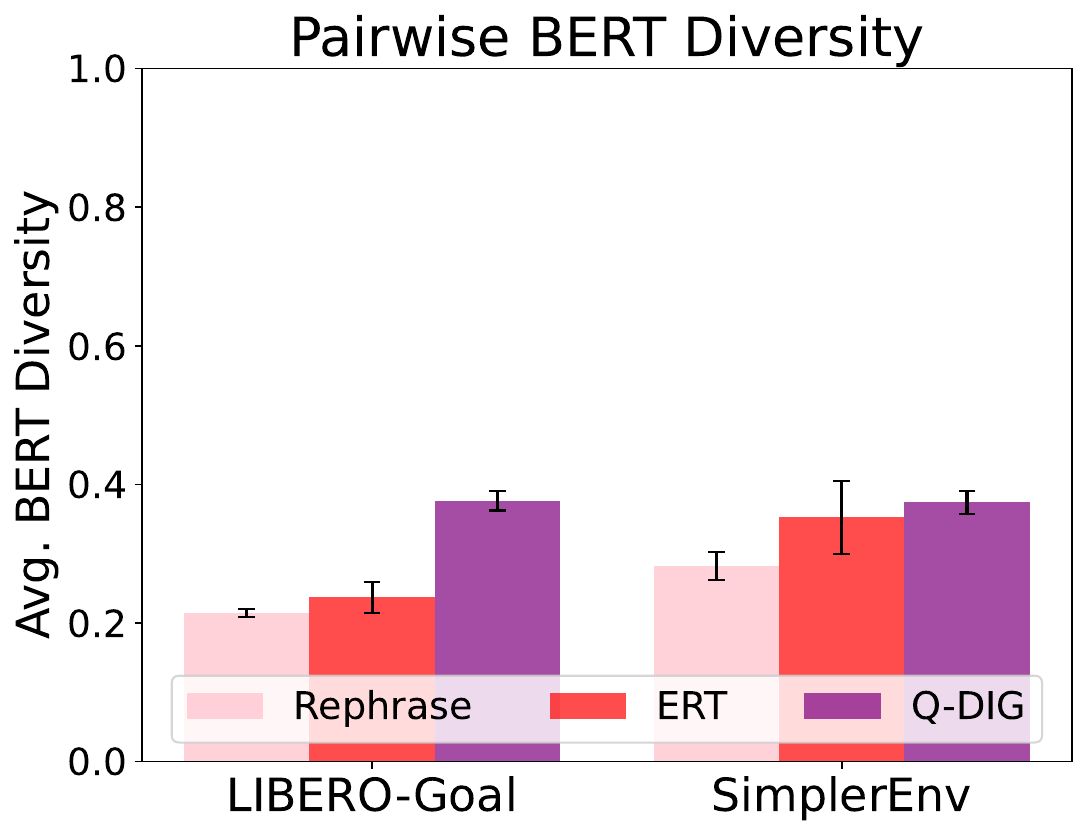} 
        \caption{BERT Diversity of prompt set.}
        \label{fig:bert_div}
    \end{subfigure}
    \hfill
    \begin{subfigure}[t]{0.325\textwidth}
        \includegraphics[width=\linewidth]{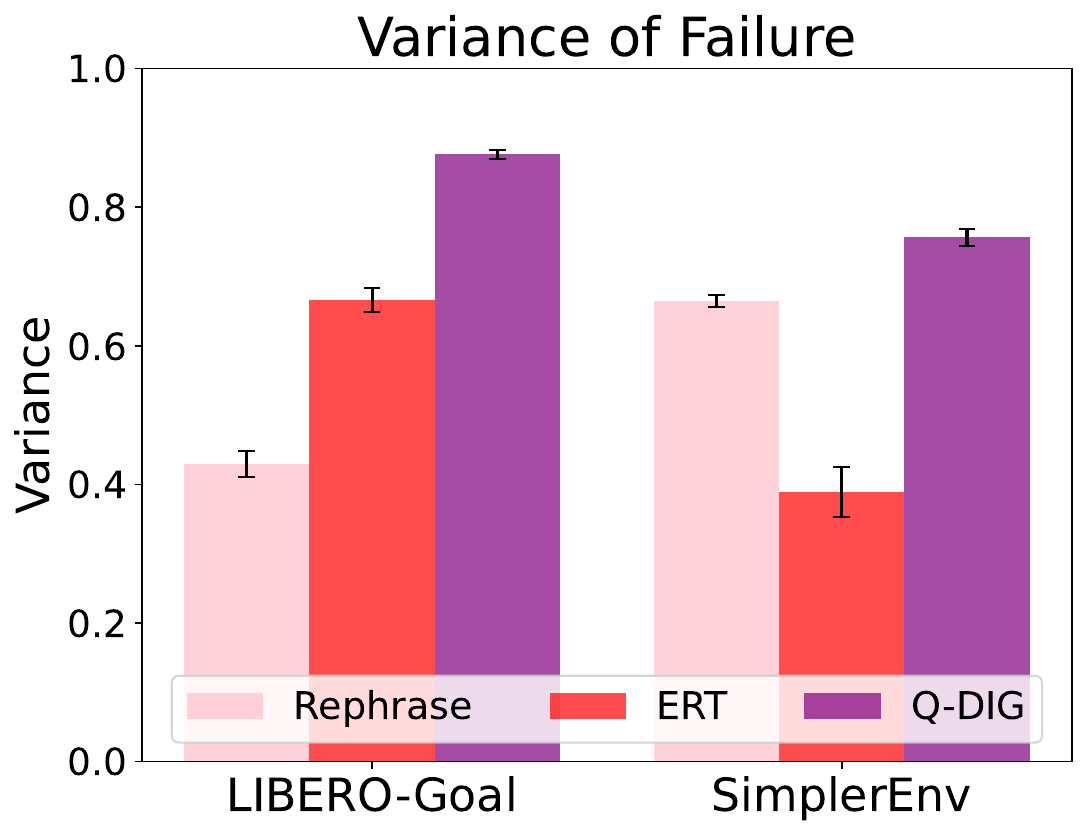} 
        \caption{Failure Variance.}
        \label{fig:failure_variance}
    \end{subfigure}
    \hfill
    \begin{subfigure}[t]{0.325\textwidth}
        \includegraphics[width=\linewidth]{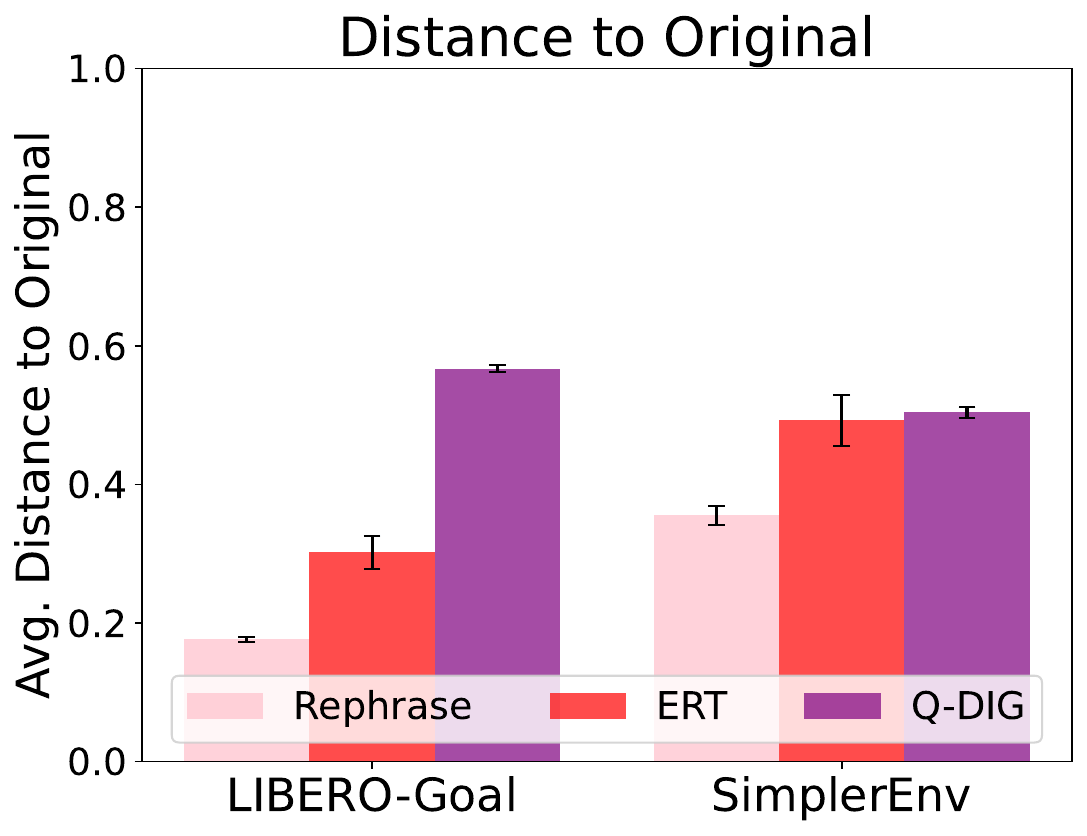}
        \caption{Distance to original prompt.}
        \label{fig:orig_div}
    \end{subfigure}

    \caption{
        Diversity of our generated data compared to the Rephrase and ERT~\cite{karnik2024embodiedredteamingauditing} baselines on OpenVLA-OFT.
        Each experiment was repeated 4 times, with error bars representing the standard error of the measurements. 
        ``Variance of Failure'' is rescaled to be between 0 and 1.
        ``Distance to Original'' represents the average sentence embedding dissimilarity (1 - cosine similarity) for each domain (see Sec.~\ref{sec:exp_inst_gen}).
        \method obtains the highest diversity metric in all cases. %
    }
    \label{fig:method_diversity}
    \vspace{-4mm}
\end{figure*}

\subsection{Generating Adversarial Instructions}
\label{sec:exp_inst_gen}

\noindent\textbf{Domain:}
We evaluate \method in SimplerEnv~\cite{li24SimplerEnv} and LIBERO~\cite{liu2023libero}, two standard simulation domains for evaluating generalist robot policies. 
We use five tasks in SimplerEnv (picking up a Coke can, an apple, and a sponge, and opening and closing the top and bottom drawers) and all ten in the LIBERO-Goal task suite (different goals from a common starting setup, e.g., turning on the stove, putting a bowl on the stove, etc.) to highlight the effect of language instructions on VLA performance. 
We tested on other suites in LIBERO (e.g., LIBERO-Spatial, where objects were placed in different spatial arrangements for different tasks) but found that VLAs overfit to visual information of those tasks.
Due to the one-to-one mapping between the starting observation and the task, VLAs solved those tasks with only image observations.

To ensure a good baseline performance of VLAs on these tasks before adversarial instruction generation, we fine-tuned them with the provided demonstration dataset for LIBERO and demonstrations collected via teleoperation for SimplerEnv. 
For both domains, we selected OpenVLA~\cite{kim24openvla} and followed the process outlined by OpenVLA-OFT~\cite{kim2025fine} for fine-tuning the base VLA for each domain. 
For LIBERO, we also conducted the same experiments on $\pi_{0.5}$~\cite{intelligence2025pi05} and GR00T~N1.6~\cite{gr00tn1_2025}. We limited these additional VLAs to LIBERO due to time and computational constraints. 

\noindent\textbf{Baseline 1, Embodied Red Teaming (ERT)~\cite{karnik2024embodiedredteamingauditing}:}
Our first baseline, ERT prompts a VLM to generate new instructions by providing the initial image observation of the task and example instructions.
Unlike \method, the in-context examples are selected purely based on whether they cause the VLA to fail, and no attack styles are targeted.
The complete loop of ERT involves (1) generating multiple sets of candidate instructions, each of size $b$, (2) selecting the set with the highest diversity (average pairwise distance between the instructions' sentence embeddings), (3) testing whether each new instruction causes the VLA to fail over multiple evaluations, and (4) selecting the instructions leading to the highest failure rates as in-context examples for the next iteration.

\noindent\textbf{Baseline 2, Rephrase:}
Following prior work~\cite{karnik2024embodiedredteamingauditing}, we chose a second baseline, Rephrase, that prompts an LLM to rephrase the original instruction into $b$ variants. 
The exact prompts are available on our \href{qdigvla.github.io}{project website}.

\noindent\textbf{Experiment Setup:}
For a fair comparison, we ran ERT and \method for the same number of iterations $T$ (3 for SimplerEnv and 6 for LIBERO, due  to differences in the base VLA performance), batch size $b=10$, and the number of generated instruction sets per iteration for rejection sampling $k=5$.
For $\pi_{0.5}$, we found that the model required more iterations to red-team, and thus doubled the total QD iterations to 12.
For \method, we chose the set of attack styles $\gZ$ as outlined in Table~\ref{tab:failure_categories}.  
Of the eight total attack styles, five were obtained from prior work~\cite{karnik2024embodiedredteamingauditing}, and three were added based on qualitative observations of prompts inducing failures. 
We chose \texttt{gpt-4o-2024-08-06} as both the VLM for ERT and \method, and as the LLM for Rephrase and the LLM judge in \method.
Prompts are available on our \href{qdigvla.github.io}{project website}. 

\noindent\textbf{Metrics:}
We compare the diversity of instructions generated by \method and the baselines with four \emph{diversity metrics}: (1) BERT diversity, (2) BLEU diversity, (3) dissimilarity to the original instruction, and (4) archive coverage.
The first two metrics, adopted from prior work~\cite{tevet2020evaluating,hong2024curiosity,karnik2024embodiedredteamingauditing}, calculate the average pairwise similarity of the BERT embeddings~\cite{reimers2019sentence} and the average pairwise BLEU score~\cite{papineni2002bleu}, respectively.
The third metric measures the average cosine similarity between the BERT embeddings of the original instructions and the generated instructions.
We report ``${1-\text{similarity}}$'' for these three diversity metrics.
The fourth is a QD metric~\cite{pugh2016quality} that maps the generated instructions to the attack styles $\gZ$ and computes the fraction of categories that have an instruction mapping to them (i.e., fraction of filled archive cells).
Besides the diversity metrics, we compared the average failure variance induced by the generated instructions as a proxy for whether the instructions are within the VLA's capability.

We aggregated metrics over four independent runs of each method, where each run performs the full instruction-generation pipeline with different stochastic LLM outputs, producing a different set of candidate instructions and corresponding diversity and failure metrics.
We tested hypothesis \textbf{H1}: \emph{\method will obtain higher diversity metrics compared to the baselines due to its targeted instruction mutations.}

\begin{figure}[t]
    \centering
    \begin{subfigure}[t]{0.46\columnwidth}
        \includegraphics[width=\linewidth]{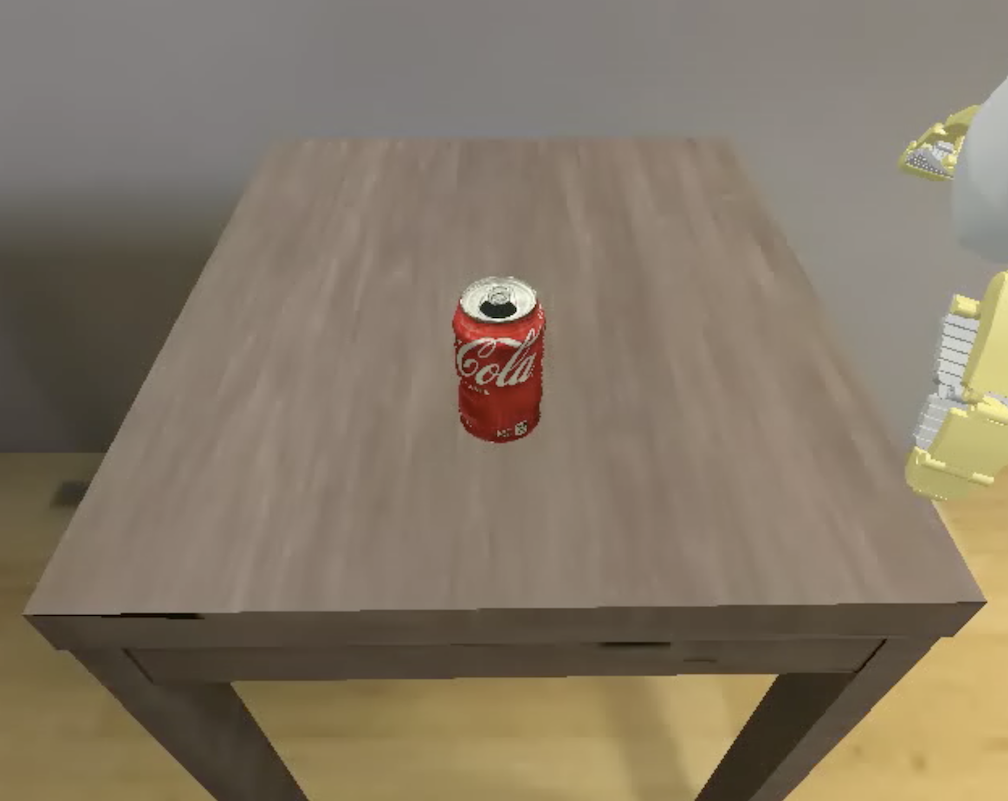} 
        \caption{Start state}
        \label{fig:user_study_start}
    \end{subfigure}
    \hfill
    \begin{subfigure}[t]{0.46\columnwidth}
        \includegraphics[width=\linewidth]{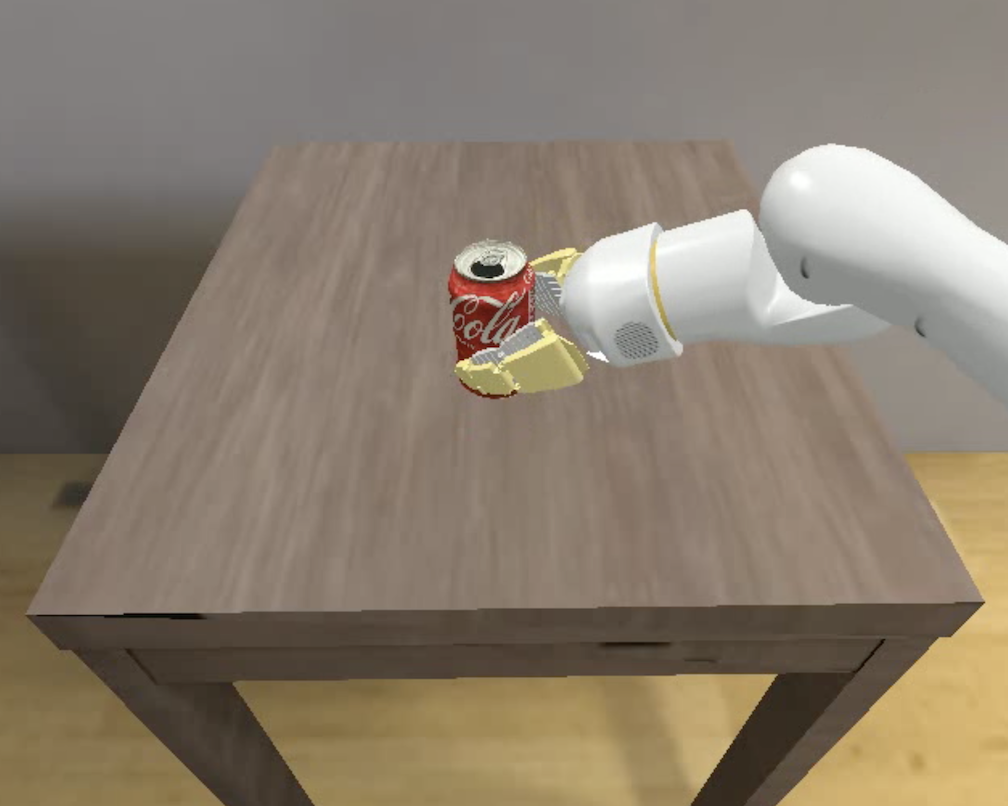} 
        \caption{Goal state}
        \label{fig:user_study_goal}
    \end{subfigure}
    \caption{
        An example of task images shown in our user study.
        Users were asked two sets of questions.
        First, they were asked to enter two instructions, one with natural wording and one attempting to be adversarial. 
        Second, they were asked to rank and rate the human-likeness of instructions generated by \method, ERT, and Rephrase.
    }
    \label{fig:user_study}
    \vspace{-5mm}
\end{figure}

\noindent\textbf{Subjective Comparison of Instructions' Human-likeness:}
We conducted a user study ($n = 40$ participants) to measure whether the generated instructions were human-like.
For each of the five tasks from SimplerEnv, we provided the initial and goal-state image (see Fig.~\ref{fig:user_study}), and first, asked the participants to write down two instructions for the robot to complete the task: one similar to how they would naturally instruct, and one that they think would confuse the robot.

We then showed participants three variations of the initial task instruction, one generated by Rephrase, ERT, and \method each (in a randomized order).
Participants were asked to rank the three instructions in decreasing order of human-likeness, and rate the human-likeness of each instruction on a 7-point Likert scale.
We tested hypothesis \textbf{H2}: \emph{Instructions generated by \method are more human-like than those generated by the baselines, since it explicitly searches for diverse attack styles.}

\subsection{VLA Fine-Tuning}
\label{sec:exp_finetuning}

\noindent\textbf{Dataset and fine-tuning:}
To test the efficacy of the generated prompts for fine-tuning VLAs, we created three augmented demonstration datasets for each domain (as described in Sec.~\ref{sec:method_finetuning}), with generated adversarial instructions from \method and the baselines.
With each algorithm, we generated 10 instructions per task, of which we randomly chose 8 for augmenting the dataset and held out 2 for evaluation.
With these augmented datasets, we followed the fine-tuning process outlined by each VLA architecture to obtain three fine-tuned VLAs corresponding to the three augmented datasets for each VLA architecture. %
We use 10,000 fine-tuning iterations for $\pi_{0.5}$ and OpenVLA-OFT, and 20,000 for GR00T N1.6.

\noindent\textbf{Evaluation:}
We evaluated the fine-tuned VLAs on all (seen and unseen) instructions for each task in LIBERO-Goal.
For each instruction, we repeated the evaluation 50 times, sampling the start state from the predefined start state distribution of each task.
We tested the following hypotheses:

\noindent\textbf{H3:}
\emph{
Fine-tuning on each augmented dataset leads to improved VLA performance compared to the base VLA, since the augmented datasets have diverse instructions.
}

\noindent\textbf{H4:}
\emph{
Fine-tuning on the dataset augmented with \method prompts leads to better VLA performance compared to fine-tuning with other augmented datasets, since we hypothesize the diversity of \method-generated instructions to be higher.
}

\subsection{Real World Experiments}
\label{sec:exp_real}

To demonstrate the effect of dataset augmentation with \method on VLA fine-tuning for real robots, we chose two representative tasks from our simulation experiments: ``push the coke can'' and ``push the sponge.'' We collected a base dataset of 50 real-world demonstrations using a Gen-2 Kinova JACO arm and two RealSense Depth Camera D435i (third person and wrist view). See Figure~\ref{fig:pull} for a visual of our real-world setup. 
To generate adversarial prompts with \method, we implemented digital twins of these tasks in SimplerEnv and ran three \method iterations to produce an archive of adversarial task instructions for evaluation.

We first fine-tuned a base OpenVLA-OFT policy on real-world demonstrations to complete tasks using the original instructions, and evaluated it on \method prompts generated in simulation by executing it on the real robot. We then fine-tuned a second OpenVLA-OFT policy on an augmented dataset pairing demonstrations with \method-generated instructions (Sec.~\ref{sec:method_finetuning}), and evaluated it on the same real-world tasks.
We tested \noindent \textbf{H5}:
\emph{Real robots perform better with a VLA fine-tuned on the augmented dataset versus one fine-tuned on the base dataset.}

\section{Results}

We discuss our results for the prompt generation and user studies (Sec.~\ref{ssec:results_prompts}), fine-tuning three VLAs (Sec.~\ref{ssec:results_finetunedVLAs}), and real-world deployment (Sec.~\ref{ssec:real_world_exp}).

\subsection{Results: Prompt Generation}
\label{ssec:results_prompts}

\begin{figure}
    \centering

    \includegraphics[width=1\linewidth]{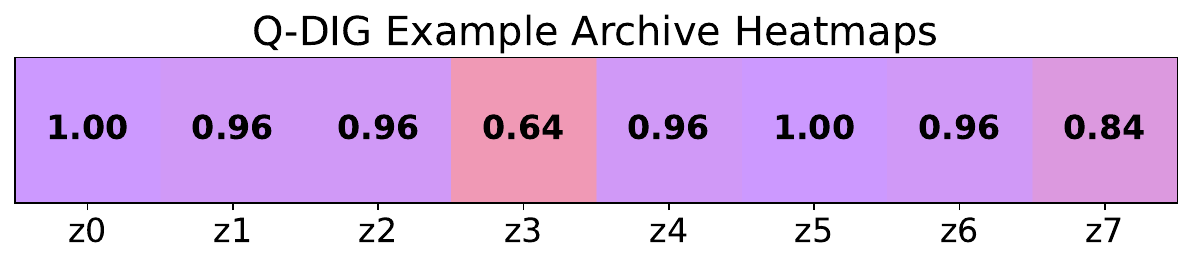}

    \caption{Example archive heatmap from \method on the LIBERO-Goal ``put the bowl on top of the cabinet'' task for OpenVLA-OFT. %
    The ``z0'' through ``z7'' labels refer to the failure modes outlined in Table~\ref{tab:failure_categories}. 
    For example, ``z0'' corresponds to ``step-by-step instructions.'' 
    The value in the cell corresponds to the failure variance of the discovered instruction, from 0 to 1.
    \method discovers instructions of diverse attack styles with high failure variance. %
    }
    \label{fig:qdvlm_heatmaps}
    \vspace{-2mm}
\end{figure}

\begin{table*}[h!]
\centering
\caption{Fine-tuned VLA results on the LIBERO-Goal task suite, using (i) original and (ii) adversarial plus original instructions across three VLAs: OpenVLA-OFT, $\pi_{0.5}$, and GR00T N1.6. 
``Original'' represents evaluation on the original task instruction.
Each instruction was evaluated 50 times.
``Rephrase,'' ``ERT,'' and ``\method'' represent evaluation on 2 unseen adversarial instructions.
Columns correspond to the original VLA and models fine-tuned with instructions from the corresponding algorithm, and rows correspond to unseen instructions during training.
Models trained on adversarial instructions (ERT and \method rows) perform better on unseen adversarial instructions.
}
\label{tab:fine_tune_vla_experiments_3vla}
\resizebox{\textwidth}{!}{
\begin{tabular}{|l|cccc|cccc|cccc|}
\hline
\multirow{2}{*}{Unseen Prompts} 
& \multicolumn{4}{c|}{\textbf{OpenVLA-OFT}} 
& \multicolumn{4}{c|}{$\mathbf{\pi_{0.5}}$}
& \multicolumn{4}{c|}{\textbf{GR00T N1.6}} \\

\cline{2-13}

& Orig & Reph & ERT & Q-DIG
& Orig & Reph & ERT & Q-DIG
& Orig & Reph & ERT & Q-DIG \\

\hline

Original 
& \textbf{97.4} & 62.2 & 94.6 & 93.8
& \textbf{96.8} & \textbf{96.8} & 96.4 & 96.4
& \textbf{77.0} & 51.6 & 46.6 & 64.8 \\

Rephrase 
& 90.3 & 63.4 & \textbf{90.8} & 84.7
& 90.7 & \textbf{96.0} & 95.9 & 95.7
& 28.2 & 44.4 & 30.1 & \textbf{47.7} \\

ERT 
& 66.2 & 53.3 & \textbf{91.1} & 66.9
& 62.3 & \textbf{72.7} & 69.6 & 70.8
& 18.1 & 36.1 & \textbf{41.9} & 37.6 \\

Q-DIG 
& 63.9 & 47.6 & 76.4 & \textbf{88.9}
& 67.5 & 71.9 & \textbf{74.8} & 73.4
& 33.7 & 42.4 & 33.6 & \textbf{55.1} \\

\hline

Avg
& 76.9 & 55.8 & \textbf{87.3} & 82.1
& 76.8 & \textbf{82.6} & 82.4 & 82.3
& 33.9 & 42.5 & 36.8 & \textbf{49.4} \\

\hline
\end{tabular}}
\vspace{-3mm}
\end{table*}

\noindent\textbf{Instructions Generated by \method obtain higher diversity metrics compared to baselines (H1):}

\begin{table}[h]
\centering
\caption{
    Archive Coverage of failure modes (Table \ref{tab:failure_categories}) and pairwise BLEU diversity results across all 3 methods and 2 benchmarks for our OpenVLA-OFT fine-tuned model.
    We show the values along with their standard errors in subscripts.
}
\label{tab:bleu_and_archive_coverage}
    \begin{tabular}{|l|c|c|c|}
    \hline
    Metric (Domain) & Rephrase & ERT & \method \\ \hline
    \textbf{BLEU (LIBERO)} & $\textbf{0.963}_{0.003}$ & $0.951_{0.01}$ & $0.947_{0.01}$ \\ \hline
    \textbf{BLEU (SimplerEnv)} & $0.928_{0.01}$ & $0.936_{0.01}$ & $\textbf{0.946}_{0.01}$ \\ \hline
    \textbf{Coverage (LIBERO)} & $0.363_{0.02}$ & $0.325_{0.01}$ & $\textbf{0.972}_{0.02}$ \\ \hline
    \textbf{Coverage (SimplerEnv)} & $0.388_{0.03}$ & $0.325_{0.02}$ & $\textbf{0.913}_{0.02}$ \\ \hline
    \end{tabular}
\vspace{-3mm}
\end{table}

We present a visual comparison of our method, \method, versus two standard baselines, Rephrase and ERT, in Figure~\ref{fig:method_diversity}.
Across our two benchmarks, we obtain higher pairwise BERT diversity (Figure~\ref{fig:bert_div}, $p<0.01$ for LIBERO-Goal) and higher or competitive dissimilarity to the original task instruction (Figure~\ref{fig:orig_div}).
Furthermore, despite the higher dissimilarity to the original instruction, our discovered instructions are within the capabilities of the VLA, as indicated by the high failure variance (Figure~\ref{fig:failure_variance}, $p < 0.01$ for both benchmarks).

In the first two rows of Table~\ref{tab:bleu_and_archive_coverage}, we report the BLEU diversity (average pairwise BLEU score), where we found a similar performance across all three methods. 
Rephrase generates semantically diverse instructions, resulting in a high average pairwise BLEU diversity, but these instructions do not cover all the realistic failure modes outlined in Table~\ref{tab:failure_categories}. %
To evaluate the archive coverage of all methods, we query the LLM judge (Sec.~\ref{ssec:gen_adv_instr}) to classify generated instructions for all three methods according to a single failure mode from Table~\ref{tab:failure_categories}.
We found that Rephrase and ERT struggled to properly target all failure categories compared to \method (last two rows of Table~\ref{tab:bleu_and_archive_coverage}).
See Figure~\ref{fig:qdvlm_heatmaps} for a heatmap of a sample instruction archive discovered by \method, highlighting the corresponding failure variance induced by the best instruction in each cell.

\noindent\textbf{Instructions Generated by \method are more human-like:}
To test \textbf{H2}, we analyzed the human-likeness rankings and ratings from our user study.
The average rankings for instructions generated by \method, ERT, and Rephrase were 1.67, 2.24, and 2.10, respectively. Hence, users consistently considered \method-generated instructions to be the most human-like.
A Kruskal-Wallis test followed by post-hoc Dunn's test with Bonferroni correction showed \method's ranking to be significantly better ($p < 0.001$) than the baselines, \textbf{validating H2}.
This was also partially supported by the Likert scores, with \method-generated instructions being rated significantly more human-like than ERT-generated instructions (mean difference of 1.07; $p < 0.001$). 
However, the difference was not significant when compared to Rephrase-generated instructions.
In conclusion, targeting specific attack styles and optimizing for failure variance resulted in more human-like instructions being generated by \method. 
While Rephrase achieves comparable human-likeness, Sec.~\ref{ssec:results_finetunedVLAs} examines how these instructions affect downstream VLA fine-tuning.

\subsection{Results: VLA Fine-Tuning}
\label{ssec:results_finetunedVLAs}

We provide results of the fine-tuned VLAs in the LIBERO-Goal task suite on OpenVLA, $\pi_{0.5}$, and GR00T N1.6. In Table~\ref{tab:fine_tune_vla_experiments_3vla}, the rows correspond to the source of unseen evaluation instructions, and columns correspond to the base VLA and VLAs fine-tuned on instructions generated by each of the three algorithms.
Our results show that augmenting the fine-tuning dataset with adversarial instructions improves performance on unseen adversarial instructions.
OpenVLA-OFT fine-tuned with adversarial instructions from both \method and ERT increases average success rates by 5--10\% and by up to 25\% on unseen instructions generated by the same algorithm.
Similarly, $\pi_{0.5}$ fine-tuned with \method achieves a 5\% average performance improvement on unseen adversarial prompts during fine-tuning, and GR00T fine-tuned with \method achieves a 15\% improvement (see the ``Avg'' row in Table~\ref{tab:fine_tune_vla_experiments_3vla}). These results \textbf{validate H3}. 

We further evaluate our \method OpenVLA-OFT fine-tuned model using 5-fold cross validation, partitioning the ten generated prompts into five groups of two, fine-tuning on eight prompts, and evaluating on the remaining two, yielding an average success rate of $87.12 \pm 3.39$ on unseen \method prompts.
The low variance across folds suggests that robustness improvements are stable across prompt splits rather than driven by a favorable subset. %
While aggregate averages vary slightly across models, fine-tuning on \method-generated instructions performs the best on unseen \method-generated instructions for OpenVLA-OFT and GR00T N1.6, which maintain the highest human-likeness according to our user study (Sec.~\ref{ssec:results_prompts}). These results partially \textbf{validate H4}. 

We hypothesize that the performance degradation when fine-tuning OpenVLA-OFT with Rephrase instructions stems from a lack of targeted, visually-grounded failure mode coverage.
Without visual grounding, Rephrase introduces linguistic variations that, while semantically diverse, may not target base VLA's multimodal vulnerabilities, leading to overfitting on irrelevant textual quirks during fine-tuning. 
In contrast, \method filters instructions through a failure variance objective, ensuring prompts remain task-relevant, human-like, and within the model's physical execution boundaries.

We additionally evaluate fine-tuned OpenVLA-OFT models in SimplerEnv across five manipulation tasks. All three augmentation strategies improve robustness to unseen adversarial instructions compared to the base VLA (37.0\%), with \method, ERT, and Rephrase achieving 52.2\%, 54.3\%, and 60.5\%, respectively. Excluding the open top drawer task, where the base VLA struggles due to confusion between semantically similar tasks, \method achieves the highest success rate (63.6\%), outperforming Rephrase (61.5\%) and ERT (57.2\%). This suggests Q-DIG is most effective when the base VLA already has sufficient task competence. %
In summary, fine-tuning with adversarially generated instructions improves robustness to unseen adversarial prompts across all three VLAs compared to the base models, suggesting that VLAs should be trained with diverse adversarial prompts.

\subsection{Results: Real-World Experiments}
\label{ssec:real_world_exp}

We evaluate our real-world policy (Sec~\ref{sec:exp_real}) on three adversarial instructions generated from \method (see Table~\ref{tab:real_world_experiments}). %
Prompts generated from \method in simulation follow similar adversarial trends in the real world to those in simulation. %
For example, Prompts 1 (P1) and 3 attain a lower success rate in simulation and the real world, and Prompt 4 attains a 100\% success rate in both simulation and the real world. %

\begin{table}[h!]
\centering
\caption{
    Comparison of Original and \method instruction variations for the ``push the coke can" real-world task.
}
\label{tab:real_world_experiments}
\begin{tabular}{|l|c|c|c|c|c|}
\hline
 & Original & P1 & P2 & P3 & P4 \\ \hline
Real-World & 9/10 & 0/10 & 8/10 & 5/10 & 10/10 \\ \hline
Simulation & 10/10 & 4/10 & 7/10 & 4/10 & 10/10 \\ \hline
\end{tabular}
\vspace{-1mm}
\end{table}

\begin{table}[h]
\centering
\caption{
    Unseen prompt evaluated on Original vs. Original and \method augmented models for ``push the coke can" real-world task.
}
\label{tab:real_world_augmented_policy_eval}
\begin{tabular}{|l|c|c|}
\hline
& Unseen P1 & Unseen P2 \\ \hline
Original & 0/10 & 8/10 \\ \hline
\method & 7/10 & 9/10 \\ \hline
\end{tabular}
\vspace{-3mm}
\end{table}

We also evaluate our data augmented policy in the real-world for ``push the coke can'' task in Table~\ref{tab:real_world_augmented_policy_eval} to validate the real data augmentation hypothesis \textbf{H5}.
We find that two unseen prompts (the same P1 and P2 in Table~\ref{tab:real_world_experiments}) perform better in the real world after performing data augmentation during fine-tuning, \textbf{validating H5}. The exact prompts are available on our \href{qdigvla.github.io}{project website}.

\section{Conclusion and Limitations}

We propose \method, a method for red-teaming robotic Vision-Language-Action models that leverages Quality Diversity and Vision-Language Models to generate diverse adversarial prompts. Compared to baselines, \method generates visually-grounded, more diverse, more human-like adversarial instructions while remaining in-distribution to expose meaningful vulnerabilities in VLAs. 
We also demonstrate that fine-tuning with diverse adversarial instructions improves the robustness of VLA policies to unseen instructions.

One limitation of \method is that it evaluates the success rate induced for a given instruction by rolling out the VLA in the environment multiple times.
Hence, since VLA rollouts are computationally expensive, we were only able to run \method for 3--12 iterations.
Another limitation is that the instructions are not generated with feedback from VLA training (i.e., leveraging fine-tuning losses), and hence, do not maximize learning potential. 
In the future, we will explore ways to reduce the computation required through approaches such as surrogate modeling~\cite{bhatt2023}, and thus enable even more scalable red-teaming of VLAs. 
Overall, we hope that our work paves the way for future work in scalable red-teaming methods for robust generalist robots.

\section*{Acknowledgments}
This work was supported by NSF CAREER \#2145077 and DARPA EMHAT HR00112490409.

\renewcommand{\bibfont}{\footnotesize}  
\printbibliography

\end{document}